\documentclass[sigconf]{acmart}

\AtBeginDocument{%
  \providecommand\BibTeX{{%
    \normalfont B\kern-0.5em{\scshape i\kern-0.25em b}\kern-0.8em\TeX}}}

\setcopyright{acmcopyright}
\copyrightyear{2023}
\acmYear{2023}
\acmDOI{XXXXXXX.XXXXXXX}

\acmConference[WWW'23]{Proceedings of the ACM Web Conference 2023}{April 30 - May 4, 2023}{Austin, Texas, USA}
\acmPrice{15.00}
\acmISBN{978-1-4503-XXXX-X/18/06}

\begin{document}

\title{Reinforcing User Retention in a Billion Scale Short Video Recommender System}

\author{Qingpeng Cai}
\affiliation{%
  \institution{Kuaishou Technology}
  \city{Beijing}
  \country{China}}
\email{caiqingpeng@kuaishou.com}

\author{Shuchang Liu}
\affiliation{%
  \institution{Kuaishou Technology}
  \city{Beijing}
  \country{China}}
\email{liushuchang@kuaishou.com}
\authornote{The first two authors contributed equally to this work}

\author{Xueliang Wang}
\affiliation{%
  \institution{Kuaishou Technology}
  \city{Beijing}
  \country{China}}
\email{wangxueliang03@kuaishou.com}

\author{Tianyou Zuo}
\affiliation{%
  \institution{Kuaishou Technology}
  \city{Beijing}
  \country{China}}
\email{zuotianyou@kuaishou.com}

\author{Wentao Xie}
\affiliation{%
  \institution{Kuaishou Technology}
  \city{Beijing}
  \country{China}}
\email{xiewentao@kuaishou.com}

\author{Bin Yang}
\affiliation{%
  \institution{Kuaishou Technology}
  \city{Beijing}
  \country{China}}
\email{yangbin11@kuaishou.com}

\author{Dong Zheng}
\affiliation{%
  \institution{Kuaishou Technology}
  \city{Beijing}
  \country{China}}
\email{zhengdong@kuaishou.com}

\author{Peng Jiang}
\affiliation{%
  \institution{Kuaishou Technology}
  \city{Beijing}
  \country{China}}
\email{jiangpeng@kuaishou.com}
\authornote{Corresponding author}

\author{Kun Gai}
\affiliation{%
  \institution{Unaffiliated}
  \city{Beijing}
  \country{China}}
\email{gai.kun@qq.com}

\renewcommand{\shortauthors}{Cai, Liu, et al.}

\begin{abstract}
Recently, short video platforms have achieved rapid user growth by recommending interesting content to users. 
The objective of the recommendation is to optimize user retention, thereby driving the growth of DAU (Daily Active Users). Retention is a long-term feedback after multiple interactions of users and the system, and it is hard to decompose retention reward to each item or a list of items. Thus traditional point-wise and list-wise models are not able to optimize retention. In this paper, we choose reinforcement learning methods to optimize the retention as they are designed to maximize the long-term performance. We formulate the problem as an infinite-horizon request-based Markov Decision Process, and our objective is to minimize the accumulated time interval of multiple sessions, which is equal to improving the app open frequency and user retention. However, current reinforcement learning algorithms
can not be directly applied in this setting due to uncertainty, bias, and long delay time incurred by the properties of user retention. We propose a novel method, dubbed RLUR, to address the aforementioned challenges. Both offline and live experiments show that RLUR can significantly improve user retention. RLUR has been fully launched in Kuaishou app for a long time, and achieves consistent performance improvement on user retention and DAU. 
\end{abstract}





\maketitle

\section{Introduction}
As newly emerging media and sharing platforms, short video applications like TikTok, YouTube Shorts, and Kuaishou quickly attract billions of users by recommending interesting content to them. 
There has been an increasing interest in short video recommendation \cite{wang2022make, zhan2022deconfounding,lin2022feature, gong2022real} in academia and industry. 
In the view of the recommender, a user interacts with the system through explicit and implicit feedback (e.g. watching time, like, follow, comment, etc.) in multiple requests, and the system returns a list of short videos at each request.
The ultimate goal is to improve retention, which reflects user satisfaction \cite{wu2017returning,zou2019reinforcement, xue2022resact}. 
User retention is
defined as the ratio of visiting the system again, and commonly referred to the retention at next day. It directly affects DAU, which is the core value of the short video app. 

Current recommender systems deploy point-wise models \cite{covington2016deep} or list-wise models \cite{pei2019personalized} to recommend items to users. 
Point-wise models predict the immediate reward of user-item pair, while list-wise models estimate the combination of rewards of a list of items considering the positions and relationships between items. 
Both of them aim at predicting a point-wise reward or a combination of point-wise rewards. 
However, retention reward is a long-term feedback after multiple interactions between users and the system. 
Similar to Go game \cite{silver2016mastering}, the relationship between retention reward and the intermediate feedback is not clear, and it is difficult to decompose retention reward to each item or a list of items. 
Thus traditional models are not able to optimize retention.

Reinforcement learning (RL) \cite{sutton1998introduction} methods are designed to learn a policy to maximize the long-term reward by interacting with the environment. Thus in this paper we choose RL methods to optimize retention.
We model the problem of optimizing user retention in short video systems as an infinite horizon request-based Markov Decision Process (MDP), where the recommender is the agent, and users serve as the environments. The session starts when the user visits the app. At each step (the user's request), the agent plays an action (the ranking weight) to ensemble scores from scoring models that predict various user feedback. The ensemble ranking function inputs both the ranking weight and the prediction scores, outputs videos with highest scores to the user. Then the user provides immediate feedback including watch time and other interactions. The session ends when the user leaves the app, the next session starts when the user opens the app again and the process repeats. Our objective is to minimize the cumulative returning time (defined as the time gap between the last request of the session and the first request of the next session), which is equal to improving the frequency of user visits, i.e. app open frequency. And the frequency of user visits directly corresponds to retention. Different from previous RL for recommender system works \cite{nemati2016optimal,zhao2017deep,  zhao2018recommendations,chen2018stabilizing, zou2019reinforcement,liu2019deep, chen2019large, xian2019reinforcement, ma2020off, afsar2021reinforcement, ge2021towards, gao2022cirs,wang2022surrogate,zhang2022multi,cai2023two, liu2023exploration, liu2023multi} that maximizes the cumulative immediate feedback, we are one of the first works to directly optimize user retention in short video recommender systems to the best of our knowledge.

However, current RL algorithms can not be trivially applied due to the
following properties of retention reward: 1) {\bf Uncertainty}: retention is not totally decided by the recommendation algorithm, and is usually affected by many uncertain factors outside the system. For example, retention is disturbed by the noise of social events. 2) {\bf Bias}: retention is biased with the different factors including time and the level of user's activity. The retention varies between weekdays and weekends, and highly active users inherently have high retention. 3) {\bf Long delay time}: different from the instant reward signal in games, the retention reward returns mostly in several hours. It causes the instability of the training of RL policy due to huge distribution shifts \cite{levine2020offline}.

For the uncertainty problem, we propose a novel normalization technique that predicts the returning time and use it to reduce the variance of the retention reward.
For the bias problem, we train different policies over user groups to prevent the learning from being dominated by high active users.
For the long delay time problem, we propose a novel soft regularization method to better balance the sample efficiency and stability.
We also take the immediate feedback as the heuristic rewards and the intrinsic motivation methods \cite{burda2018exploration,pathak2017curiosity} to better optimize user retention in the delayed reward setting. Combining the above techniques, we propose the Reinforcement Learning for User Retention algorithm, named as RLUR. 

We summarize our contributions as below: 

\begin{itemize}
\item{
We model the user retention for short video recommendation problem as an infinite horizon requested-based MDP, and the aim is to minimize the cumulative returning time.}
\item{We propose novel methods to solve the challenges of uncertainty, bias, and long delay time of user retention.}
\item{Experiments on both offline and live environments show that RLUR improves user retention significantly.}
\item{The RLUR algorithm has been fully launched in Kuaishou app, and it shows that RLUR continuously improves user retention and DAU.}
\end{itemize}

\begin{figure*}[h]
    \Description{}
    \centering
    \includegraphics[width=\textwidth]{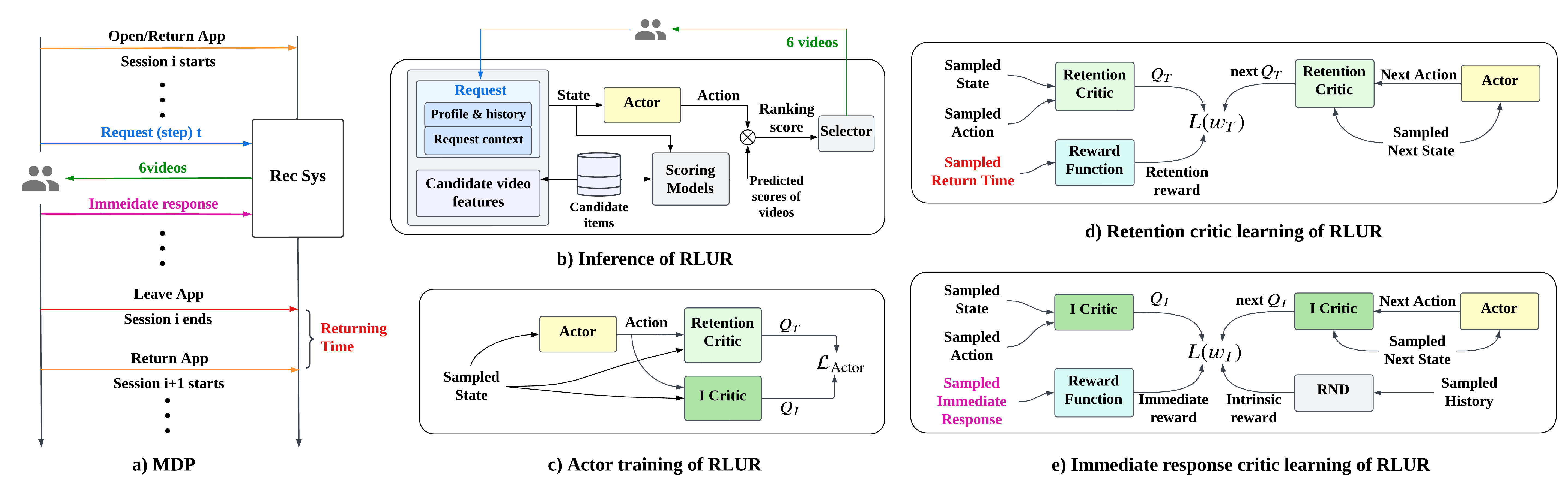}
    \caption{The infinite horizon request-based MDP and the framework of RLUR}
    \label{framework}
\end{figure*}

\section{Problem Definition}
We model the problem as an infinite horizon request-based Markov Decision Process (MDP), where the recommender is the agent, and users are the environments.
As shown in Figure \ref{framework}(a), when the user opens the app, a session $i$ starts. At each request (step) $i_t$ of the session $i$, the agent plays an action $a_{i_t}$ given the state $s_{i_t}$ of the user. $n$ deep models predict
scores $x_j=(x_{j1},...,x_{jn})$ for each candidate video $j$, in terms of various feedback(watch time, follow, like, etc). The ranking function $f$ inputs the action $a_{i_t}$ and the prediction scores $x_j$, and outputs the ranking score $f(a_{i_t},x_j)$ for each video $j$. 
The system recommends top 6 videos with the highest ranking score to the user as shown in Figure \ref{framework}(b). Then the user provides immediate feedback $I(s_{i_t},a_{i_t})$.
The session ends when the user leaves the app. 
When the user returns to the app again, the next session $i+1$ starts, and the delayed reward(returning time) of session $i$ is returned. Then the above process repeats.

We now introduce the detail of the MDP. The {\bf state} $s_{it}$ consists of user profile, behavior history $u_{i_t}$, request context, and candidate video features. The {\bf action} is a continuous vector in $[0,C]^n$, where $n$ is the number of scoring models. We use a linear {\bf ranking function}, i.e, $f(a_{i_t},x_j)=\sum_{k=1}^{n}a_{i_tk}x_{jk}$. 
The {\bf immediate reward} $I(s_{i_t},a_{i_t})$ is defined as the sum of watch time and the number of interactions at this request. The {\bf returning time} $T(s_i)$ is the time gap between the last request of session $s_i$ and the first request of session $s_{i+1}$. The {\bf returning time reward} $r(s_{i_t},a_{i_t})$ is $T(s_i)$ for the last request, and $0$ otherwise.
We learn a deterministic policy $\pi(s_{i_t}|\theta)$ that inputs the state $s_{i_t}$ and outputs the action $a_{i_t}$. The objective is to minimize the cumulative returning time $\sum_{i=1}^{\infty}\gamma^{i-1}T(s_i)$, where $\gamma(0< \gamma <1)$ is the discount factor.

\section{Method}
In this section we firstly discuss the learning of retention in the delayed reward setting. Then we propose techniques to tackle these challenges incurred by the characteristic of user retention.

\subsection{Retention Critic Learning}
In this section we discuss how to learn the retention. 
We learn a critic function $Q_{T}(s_{i_t},a_{i_t}|{w_T})$ parameterized by $w_T$, 
to estimate the cumulative returning time from the current state $s_{i_t}$ and action $a_{i_t}$. We follow the deep deterministic policy gradient method \cite{lillicrap2015continuous} to estimate the cumulative returning time reward. As shown in Figure \ref{framework}(d), the loss function $L(w_T)$ for learning returning time is 
\begin{equation}
\label{td-loss}
\sum_{s_{i_t},a_{i_t}\in D}(Q_{T}(s_{i_t},a_{i_t}|{w_T})-(r(s_{i_t},a_{i_t})+\gamma_{i_t} Q_{T}(s_{i_{t+1}},\pi(s_{i_{t+1}}|\theta)|{w_T})))^2,
\end{equation}
where the discount factor $\gamma_{i_t}$ is 1 for samples that are not the last requests and $\gamma$ for the last requests, and $D$ is the dataset. The discount factor of non-terminal samples is set as 1 to prevent the returning time reward from vanishing by the exponential decay mechanism. 
If one set the discount factor to be a number less than 1, the importance of returning time in future sessions will be extremely small. Optimizing the loss (\ref{td-loss}) is equivalent to estimate $\sum_{i=1}^{\infty}\gamma^{i-1}T(s_i)$. We omit the proof due to lack of space.

\subsection{Methods for Delayed Reward}
The returning time reward only occurs at the end of each session, and it is inefficient to learn an RL policy with delayed rewards. For better learning of user retention, we adopt the 
heuristic reward methods \cite{grzes2017reward} to enhance the policy learning and the intrinsic motivation methods \cite{burda2018exploration,pathak2017curiosity} to drive the policy to explore novel states.
As the feedback and returning time are positively correlated, we take the immediate rewards as heuristic rewards to guide the policy to improve the user retention. For better exploration, 
we choose the Random Network Distillation (RND) \cite{burda2018exploration} method that is both effective and computationally efficient. The idea is that we randomly initialize two networks with the same structure, and train one network to fit the output of another fixed network. 
The loss of RND is $L(w_e)=\sum_{{u{_{i_t}}}\in D}||E(u_{i_t}|w_{e})-E({u_{i_t}}|w_{e}^*)||_2^2$, where $E$ is the embedding function of behavior history $u_{i_t}$, $w_{e}$ is the parameter, and $w_{e}^*$ is the fixed parameter.
Intuitively, the loss of each state decreases with more samples. Thus it can quantify the novelty of each state. We use the RND loss of each sample 
as the intrinsic reward. 
To reduce the interference of immediate rewards with the retention reward, we learn a separate critic function, the immediate reward critic $Q_{I}(s_{i_t},a_{i_t}|{w_I})$ to estimate the sum of the intrinsic rewards with the immediate rewards, $r_I(s_{i_t},a_{i_t})=I(s_{i_t},a_{i_t})+||E(u_{i_t}|w_{e})-E({u_{i_t}}|w_{e}^*)||_2^2$. The loss of the immediate reward critic, $L(w_I)$ is similar to Eq.(\ref{td-loss}), shown in Figure \ref{framework}(e).

\subsection{Uncertainty}
The returning time is highly uncertain and affected by many factors outside the system. To reduce its variance , we now propose a normalization technique. We use the ratio of the true returning time over the predicted returning time as the normalized retention reward. For predicting the returning time, we learn a session-level classification model $T'$ to predict the returning probability.
We firstly calculate the empirical distribution of returning time, then use the $\beta\%$ percentile of the distribution, $T_{\beta}$ to determine whether the sample $x$ is positive or negative. The label $y$ of sample $x$ is positive if 
the returning time is less than $T_{\beta}$ as the shorter time means the better.
The loss function of $T'$ is $y*logT'(x)+(1-y)*log(1-T'(x))$, and $T'(x)$ predicts the probability that the returning time is shorter than $T_{\beta}$.
Then we get a lower bound of the expected returning time by the Markov Inequality \cite{wilhelmsen1974markov}, $ET(x)\geq P(T(x)\geq T_{\beta})*T_{\beta}\sim (1-T'(x))*T_{\beta}$ as the predicted returning time. Thus we get the normalized retention reward, $r(s_{i_t},a_{i_t})=clip\{0,\frac{T(s_i)}{(1-T'(x))*T_{\beta}},\alpha\}$, where $\alpha$ is a positive constant. 

\subsection{Bias}
As the returning time of high active users is much shorter than that of low active users, and the habit of these groups are quite different, we learn two separate policies $\pi(\cdot |\theta_\text{high})$ and $\pi(\cdot|\theta_\text{low})$ for high active group and low active group respectively. As we want to minimize the returning time and maximize the immediate rewards,
the loss of the policy $\pi(\cdot |\theta_\text{high})$
is a weighted sum of the immediate critic and retention critic, 

\begin{equation}
\begin{split}
\label{actor-loss}
L(\theta_\text{high})&=\lambda_TQ_{T}(s_{i_{t}},\pi(s_{i_{t}}|\theta_\text{high})|w_T)-\lambda_IQ_{I}(s_{i_{t}},\pi(s_{i_{t}}|\theta_\text{high})|w_I),
\end{split}
\end{equation}

where $\lambda_T, \lambda_I$ are positive weights. The loss of the other policy is similar. The learning of Actor is shown in Figure \ref{framework}(c).

\subsection{Tackling the Unstable Training Problem}
Different from that the reward returns instantly in games, the retention reward returns
in several hours to days, which causes a much larger distribution shift between current policy and behavior policy and the instability of the training of RL algorithms in our setting. Regularization methods \cite{fujimoto2021minimalist} that adds a behavior cloning loss are proposed to stabilize the training in the off-policy setting. However, we find that this method either fails to stabilize or limits the sample efficiency. To better balance the sample efficiency and the stability, we now propose a novel soft regularization method. 
The actor loss is defined as 
$\exp({\max\{\lambda* (log(p(a_{i_{t}}|s_{i_{t}}))-log(p_b(a_{i_{t}}|s_{i_{t}}))), 0\}})L(\theta),$
where $\lambda>0$ is the regularization coefficient, $p(a_{i_{t}}|s_{i_{t}})$ is the probability density of the Gaussian distribution of the current policy,
and $p_b(a_{i_{t}}|s_{i_{t}})$ is the probability density of the Gaussian distribution of the behavior policy.
The intuition is that samples with higher distribution shift get less weight in the learning,  which softly regularizes the learning.
$\lambda$ controls the regularization degree, larger $\lambda$ makes a stronger regularization on the actor loss. Note that when $\lambda$ is 0, we do not make any regularization on the actor loss.

\section{Offline Experiments}
We validate the effectiveness of RLUR on a public short video recommendation dataset, \emph{KuaiRand} ~\citep{gao2022kuairand}, which is an unbiased dataset that contains the logs of the interactions of users and the system with multiple sessions. 
We build a simulator based on this dataset, which includes three parts: 
the user immediate feedback module that predicts multiple user behaviors;
the leave module that predicts whether the user leaves the session or not; 
and the return module that predicts the probability of returning to the app on day $k \in \{1,\dots,K\}$ after each session. 
We set $K=10$.

We compare RLUR with a black-box optimization method, the Cross Entropy Method (CEM) \cite{rubinstein2004cross} that is commonly used for ranking parameter searching and a state of the art reinforcement learning method, TD3 \cite{fujimoto2018addressing}. 
We evaluate the performance of each algorithm in terms of the averaged returning day (returning time) and averaged retention on the 1st day (user retention) across all user sessions. 
We train each algorithm until convergence, and we report the averaged performance of the last 50 episodes in Table \ref{tab: offline}. 
For both metrics, TD3 outperforms CEM, which demonstrates the effectiveness of reinforcement learning. 
RLUR outperforms both TD3 and CEM significantly.
We also train a variant of RLUR that only contains the part of learning the returning time in Section 3.1, called RLUR (naive). 
RLUR (naive, $\gamma=0.9$) outperforms RLUR (naive, $\gamma=0$), which shows that it is more reasonable to minimize the cumulative retention of multiple sessions than one session as $\gamma$ controls the importance of future sessions. 
RLUR outperforms RLUR (naive, $\gamma=0.9$) substantially, which validates the effectiveness of the proposed techniques for the retention challenge.

\begin{table}[t]
    \centering
    \caption{Offline Results}
    \begin{tabular}{ccc}
        \toprule
        Algorithm &Returning time$\downarrow$& User retention$\uparrow$\\ 
        \midrule
        CEM & 2.036&0.587\\
       
        TD3& 2.009&0.592\\
         
        RLUR (naive, $\gamma=0$) &2.001&0.596 \\ 
        
        RLUR (naive, $\gamma=0.9$) & 1.961 & 0.601\\ 
        
        RLUR& {\bf 1.892}&{\bf 0.618}\\
        \bottomrule
    \end{tabular}
    \label{tab: offline}
\end{table}

\section{Live Experiments}
We test our proposed RLUR algorithm by live experiments in Kuaishou, a billion-scale short video platform. We compare RLUR with a black-box optimization baseline that is widely used for ranking parameter searching in recommender systems, CEM \cite{rubinstein2004cross}. 
We do not compare TD3 here as the training of TD3 is unstable, which is discussed in Section 3.5.
We randomly split users into two buckets, deploy RLUR in the test bucket, and run CEM in the base bucket. We evaluate the performance of algorithms in terms of app open frequency, DAU, user retention at 1st day, and user retention at 7th day. We test algorithms for a long time to get convincing results as DAU and user retention varies between days. Now we introduce the details.

    {\bf Inference and Training.} The inference procedure  is shown in Figure \ref{framework}(b). At each user request, the user state is sent to the Actor, and the mean $\mu$ and variance $\sigma$ are returned. Then the action is sampled from a Gaussian distribution $N(\mu,\sigma)$. Then the ranking function calculates the linear product of the action with the predicted scores of each video, and recommends 6 videos with the highest scores to the user. The training of actor is illustrated in Figure \ref{framework}(c). The actor is trained to minimize the weighted sum of the retention critic and the immediate reward critic, as in Eq.(\ref{actor-loss}).
    
    
    {\bf MDP.} The state is a vector of user profile, behavior history(user statistics, the id of videos and corresponding feedback of the user in previous 3 requests), request context, and the candidate video features. The user profile covers age, gender, and location. The user statistics include statistics of various feedback. For the action, we choose an 8-dimensional continuous vector ranging in $[0,4]^8$, and the policy outputs the parameters of eight scoring models  predicting the main feedback (watchtime, shortview, longview, like, follow, forward, comment, and personal-page entering). The immediate reward $I(s_{it},a_{it})$ of each request is designed as the the sum of watch time (in seconds) and interactions (including like, follow, comment, and share) of 6 videos.

    {\bf Hyper-parameters.}  The discounted factor is $0.95$, and $0.95$ outperforms other values in our system. The weights in actor loss is set to be $1.0,1.0$. As for the session-level retention model, we choose $60\%$ percentile to determine the label, and the value of upper bound, $\alpha$ is 3. The regularization coefficient $\lambda$ is $1.5.$

\begin{figure}
 \Description{}
\centering
\includegraphics[width=1.0\linewidth]{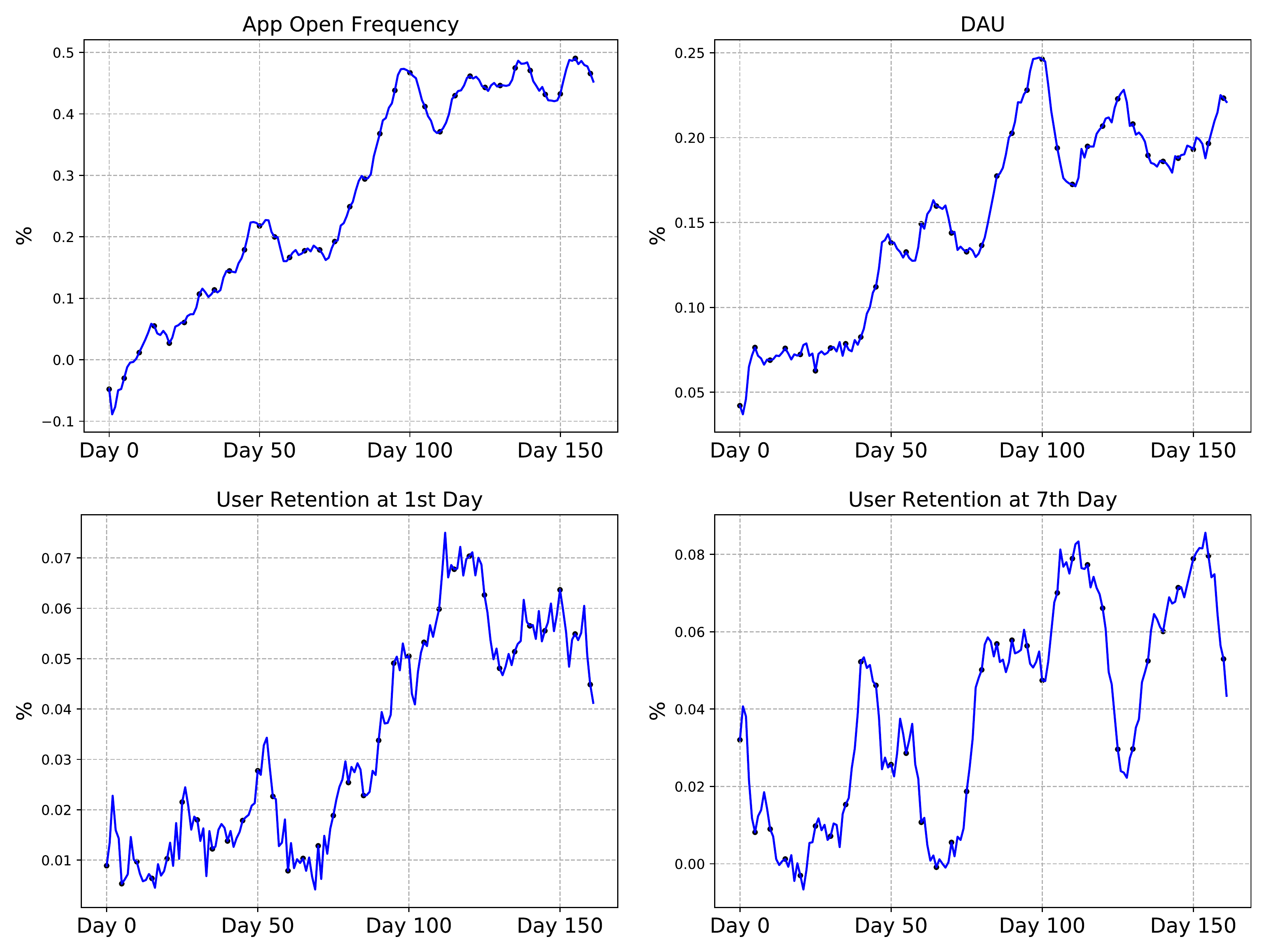}
\caption{Live performance gap of each day.}
\label{live}
\end{figure}

Figure \ref{live} plot the comparison results of RLUR with CEM, where the x axis stands for the number of days after deployment, and the y axis shows the percentage performance gap between RLUR and CEM. 
As we can see, app open frequency which directly reflects the returning time, increases consistently from Day 0 to Day 80 and sharply from Day 80 to Day 100. 
The app open frequency converges to $0.450\%$ after Day 100. That validates the training of RLUR can consistently increase the retention reward and converges at expected. 
The user retention at 1st day/7th day and DAU increase slowly with the training from Day 0 to Day 80. From day 80 to day 100, these metrics increases sharply. After day 100, the performance gap of DAU converges to $0.2\%$, user retention at 1st day converges to $0.053\%$, and user retention at 7th day converges to $0.063\%$. Note that $0.01\%$ improvement of user retention and $0.1\%$ improvement of DAU are statistically significant in short video platforms. That is, the performance of RLUR is quite significant. Note that DAU and retention continue to increase.

\section{Conclusion}
We study to optimize user retention in short video recommender systems. We discuss the challenges of optimizing user retention. We formalize the problem as an infinite-horizon request-based Markov Decision Process, and our aim is to learn a policy that minimizes the accumulated returning time. We propose novel techniques to tackle the challenge of retention, and we present the RLUR algorithm. Both offline and live experiments validate the effectiveness of RLUR. We have deployed RLUR in a billion-scale short video system for  a long time, and it improves user retention and DAU significantly and consistently.

\bibliographystyle{ACM-Reference-Format}
\bibliography{cameraready}


\end{document}